\documentclass{article}
\usepackage{spconf,amsmath,graphicx}
\usepackage{booktabs}
\usepackage{subcaption}
\usepackage{amssymb}
\usepackage{comment}

\usepackage{algpseudocode}
\usepackage[ruled,vlined]{algorithm2e}
\newcommand{\set}[1]{\mathcal{#1}}

\makeatletter
\DeclareRobustCommand\onedot{\futurelet\@let@token\@onedot}
\def\@onedot{\ifx\@let@token.\else.\null\fi\xspace}

\def\eg{\emph{e.g}\onedot} 
\def\ie{\emph{i.e}\onedot}

\def\etal{\emph{et al}\onedot}
\makeatother

\newcommand{\myparagraph}[1]{\noindent \textbf{#1}}

\def\L{{\cal L}}

\title{Box-Level Class-Balanced Sampling for Active Object Detection}

\name{Jingyi Liao, Xun Xu, Chuan-Sheng Foo and Lile Cai \thanks{This research is supported by the Agency for Science, Technology and Research (A*STAR) under its Career Development Fund (Grant no. C210812052).}}
\address{Institute for Infocomm Research (I$^2$R), A*STAR, Singapore.\\
\{liao\_jingyi, xu\_xun, foo\_chuan\_sheng, caill\}@i2r.a-star.edu.sg}

\begin{document}
%
\maketitle
\begin{abstract}
Training deep object detectors demands expensive bounding box annotation. Active learning (AL) is a promising technique to alleviate the annotation burden.  Performing AL at box-level for object detection, \ie, selecting the most informative boxes to label and supplementing the sparsely-labelled image with pseudo labels, has been shown to be more cost-effective than selecting and labelling the entire image. In box-level AL for object detection, we observe that models at early stage can only perform well on majority classes, making the pseudo labels severely class-imbalanced. We propose a class-balanced sampling strategy to select more objects from minority classes for labelling, so as to make the final training data, \ie, ground truth labels obtained by AL and pseudo labels, more class-balanced to train a better model. We also propose a task-aware soft pseudo labelling strategy to increase the accuracy of pseudo labels. We evaluate our method on public benchmarking datasets and show that our method achieves state-of-the-art performance.
\end{abstract}
\section{Introduction}
Training deep object detectors demands large amounts of bounding box annotation, which is much more expensive than assigning a class label to each image for the image classification task \cite{su2012crowdsourcing}. Active learning (AL) has emerged as a promising technique to  alleviate the annotation burden. It has been extensively studied for various visual tasks \cite{ren2021survey} and shown to obtain comparable performance of the fully-supervised baseline with significantly reduced annotated data.

The majority of previous works on active learning for object detection (ALOD) adopt an image-level approach \cite{roy2018deep,yuan2021multiple,yoo2019learning,agarwal2020contextual,wu2022entropy,yu2022consistency}, \ie, an image is treated as a sample for selection and the entire image is annotated if selected; the annotation cost is also measured by the number of annotated images. However, as the number of objects within each image can vary significantly and an image can contain multiple similar objects, an image-level approach may not reflect actual annotation cost and can waste annotation budget on labelling repetitive objects. This motivates the study of box-level approach, where a box is considered as a sample and the annotation cost is measured by the number of annotated boxes. 

Box-level approach \cite{desai2020towards,lyu2023box} is much less explored compared to image-level approach. Previous works select the most uncertain boxes for annotation. As a result of box-level selection and annotation, an image can be partially-labelled. Treating the objects in unlabelled region as background gives incorrect supervision signal and harms model performance. To address this problem, high confident predictions given by current model can be used as pseudo labels to supplement the ground truth (GT) labels during model training. 

\begin{figure*}[ht]
\centering
\begin{subfigure}{0.23\textwidth}
	\includegraphics[width=\linewidth]{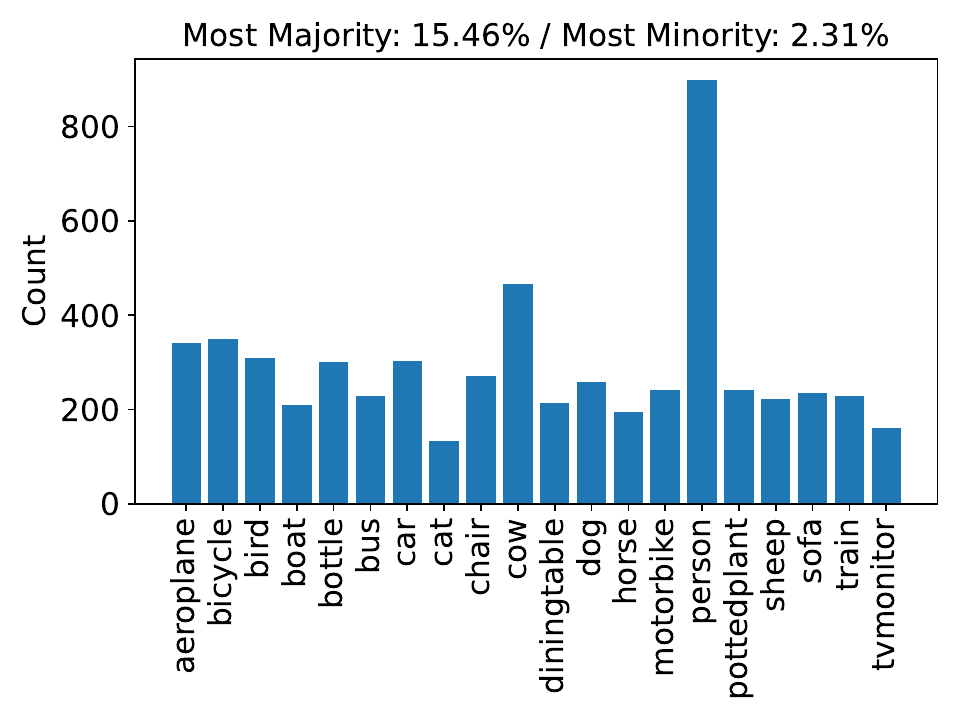}
	\caption{}
	
\end{subfigure}
\begin{subfigure}{0.23\textwidth}
	\includegraphics[width=\linewidth]{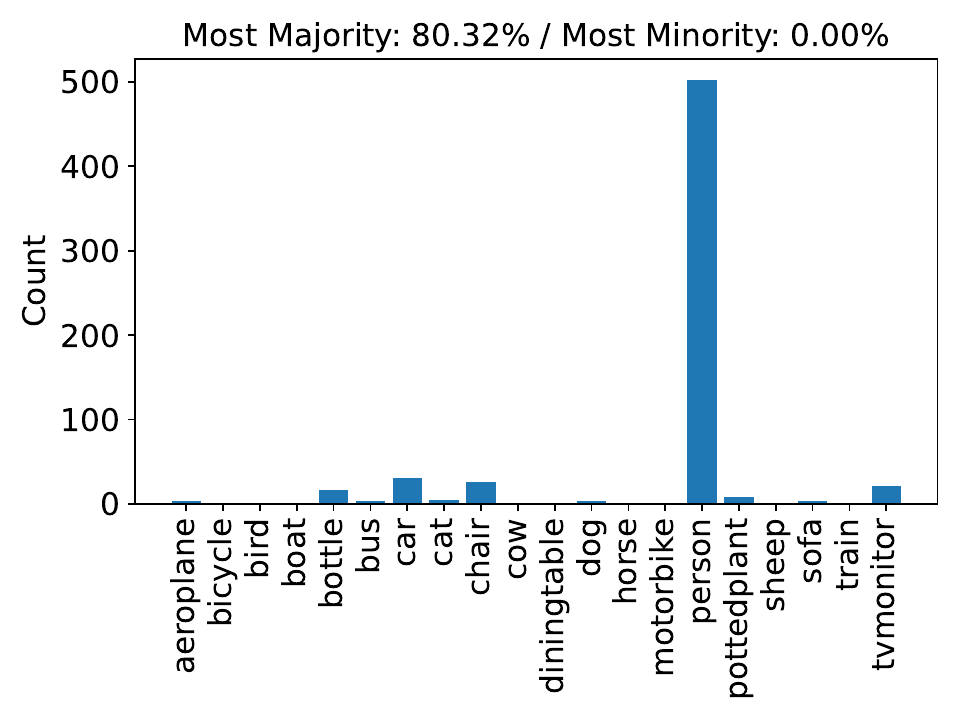}
	\caption{}
	
\end{subfigure}
\begin{subfigure}{0.23\textwidth}
	\includegraphics[width=\linewidth]{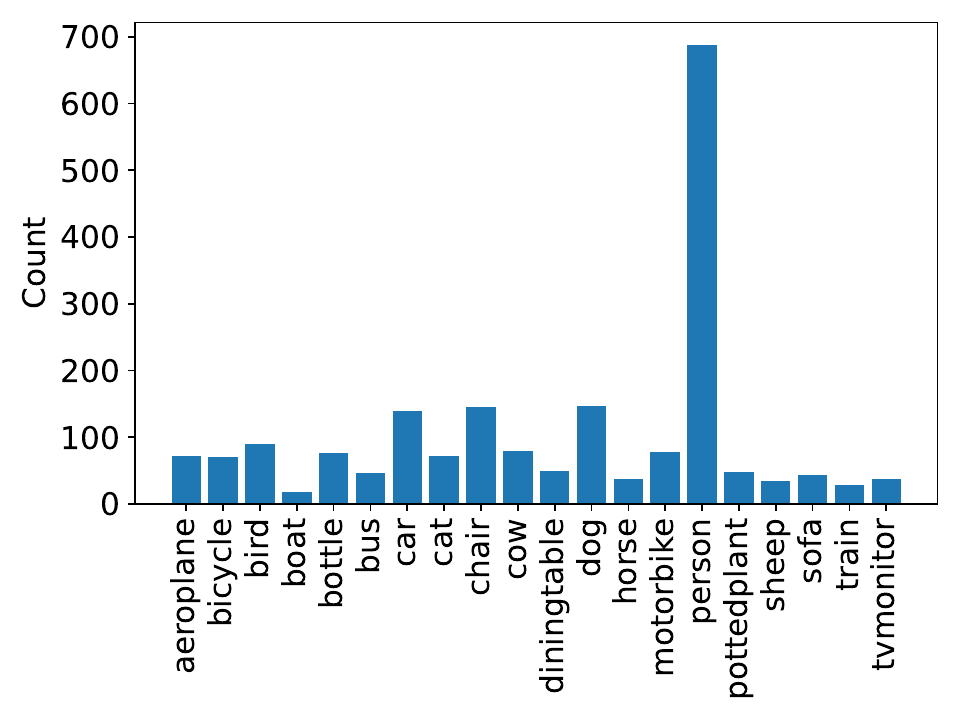}
	\caption{}
	
\end{subfigure}
\begin{subfigure}{0.23\textwidth}
	\includegraphics[width=\linewidth]{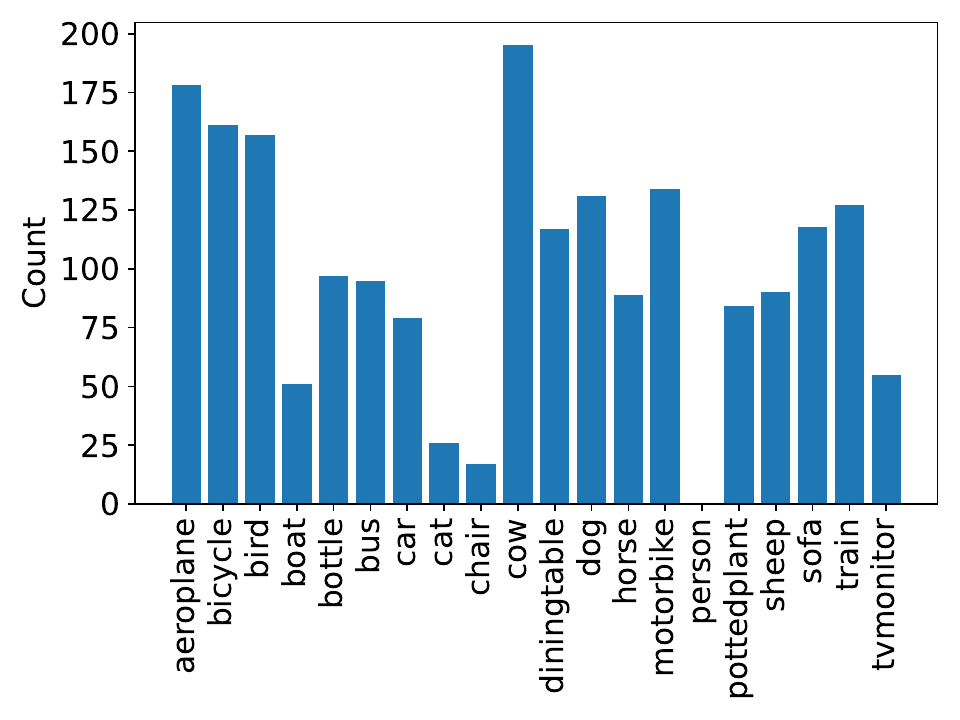}
	\caption{}
	
\end{subfigure}
\caption{(a) Class distribution of GT labels at the first batch (1000 randomly selected images). (b) Class distribution of pseudo labels (high confidence detection predicted by the model trained at the first batch). The pseudo labels have an imbalance factor even higher than GT labels. (c) Class distribution of objects selected by AL without considering class-imbalance. The selected samples are dominated by the majority class (person). (d) Class distribution of objects selected by the proposed class-balanced sampling strategy. Our method favors the selection of objects from minority classes, making the final training data, \ie, GT labels obtained by AL and pseudo labels, more class-balanced to train a better model.}
\label{fig:motivation}
\end{figure*}

An overlooked factor in previous box-level methods is the class distribution of objects. Object detection datasets typically exhibit severe class-imbalance (Fig.~\ref{fig:motivation}(a)). Models at early stage AL are trained on small amount of samples and only perform well on common objects. This has two implications. First, the pseudo labels used for sparsely labelled images are dominated by majority classes. As shown in Fig.~\ref{fig:motivation}(b), the pseudo labels have an imbalance factor even higher than GT labels, exacerbating the class imbalance problem already existing in GT labels. Second, the candidate pool for AL selection is also dominated by majority classes, resulting in more samples being selected and labelled for majority classes (Fig.~\ref{fig:motivation}(c)). These two factors combined lead to a highly imbalanced training set and degrade model performance.


Motivated by the above observation, we propose a class-balanced sampling strategy for box-level ALOD. Our strategy selects more objects from minority classes for labelling (Fig.~\ref{fig:motivation}(d)), so as to make the final training data, \ie, GT labels obtained by AL + pseudo labels, more class-balanced. Also, previous methods \cite{desai2020towards,lyu2023box} use predictions of high confidence scores (\eg, $>0.9$) as pseudo labels. However, predictions of lower confidence score can still indicate the presence of foreground objects and treating them as background will incur incorrect supervision signal. We propose a task-aware soft pseudo labelling strategy to address this problem. 

Our contributions can be summarized as below:
\begin{itemize}
    \item We identify the class-imbalance problem in box-level ALOD and propose a class-balanced AL strategy to address it. 
    \item We propose a task-aware soft pseudo labelling strategy to obtain more accurate pseudo labels for partially-labelled images.
    \item We evaluate the proposed method on public benchmarking datasets and show that our method achieves state-of-the-art performance.
\end{itemize}

\section{Related Work}
\subsection{Deep Active Learning}
As a promising technique to alleviate the annotation burden for training deep models, active learning has been extensively studied. According to the criterion used to select samples, existing methods can be categorized into uncertainty-based, diversity-based and hybrid methods. Uncertainty-based methods select the most uncertain samples for the current model to label. BALD \cite{gal2017deep} measures uncertainty by mutual information between samples and model weights. Beluch \etal \cite{beluch2018power} showed that uncertainties estimated from model ensembles perform better than Monte-Carlo Dropout uncertainties. Yoo and  Kweon \cite{yoo2019learning} proposed to learn sample uncertainty by a loss prediction module. ISAL \cite{liu2021influence} selects samples that have the most positive influence on model parameters. Diversity-based methods select a set of samples that best represent the training data distribution. CoreSet \cite{sener2017active} selects samples such that the average training loss of the selected samples approximates the average empirical loss over the entire dataset. Caramalau \etal \cite{caramalau2021sequential} proposed to perform CoreSet on features learnt from a task-aware graph convolutional network. VAAL \cite{sinha2019variational} employs adversarial training to learns a latent space to facilitate the selection of samples that are most different from labelled ones. Hybrid methods consider both uncertainty and diversity in sample selection. Yang \etal \cite{yang2017suggestive} first obtain samples of largest uncertainty, followed by diversity sampling to remove redundancy. USDM \cite{yang2015multi} formulates the selection as a quadratic programming problem, where the unary term is optimized for uncertainty and the pairwise term accounts for diversity. BADGE \cite{ash2019deep} performs diversity sampling on uncertainty-aware gradient embeddings to select a set of both uncertain and diverse samples. 

\subsection{Active Learning for Object Detection}
Depending on the granularity used to select samples for annotation, active learning methods for object detection can be divided into image-level and box-level methods. The former select an entire image to label, while the latter select part of an image for labelling. 

\myparagraph{Image-level methods} can be grouped into architecture-specific \cite{roy2018deep,yuan2021multiple,yoo2019learning} and architecture-agnostic approaches \cite{agarwal2020contextual,wu2022entropy,yu2022consistency}. Architecture-specific approach relies on the output of detector-specific layers \cite{roy2018deep} or have customized layers on top of the original detector \cite{yuan2021multiple,yoo2019learning}. Roy \etal \cite{roy2018deep} compute the margin of predictions made by different layers of one-stage object detector to measure uncertainty. MI-AOD \cite{yuan2021multiple} utilizes multiple instance learning to suppress noisy detections and  measures uncertainty by the discrepancy of two adversarial classification heads. LLAL \cite{yoo2019learning} adds a loss prediction module to learn the prediction loss of the detection model. Architecture-agnostic approach does not rely on specific architecture design. CDAL \cite{agarwal2020contextual} employs the predicted probability distribution of each box to select contextual diverse samples. EnmsDivproto \cite{wu2022entropy} excludes similar boxes when generating image-level uncertainty score and rejects redundant samples based on class-wise box level similarity. CALD \cite{yu2022consistency} exploits the prediction consistency between original and augmented images to measures the uncertainty of a sample. 

\myparagraph{Box-level methods} are much less explored compared to image-level methods. The pioneering work \cite{desai2020towards} explored uncertainty-based and diversity-based sampling at box level and showed that box-level selection is more cost-effective than image-level selection. ComPAS \cite{lyu2023box} employs a ``query-by-committee" method to measure classification and localization uncertainty for each box, and supplements sparsely labelled images with pseudo labels to alleviate the missing label problem. Previous works do not take into consideration that the pseudo labels are dominated by majority classes. In this work, we accommodate this characteristic to design a better active learning strategy for box-level selection.

\subsection{Class-Balanced Active Learning}
The class-imbalance problem in training data has been extensively studied. Loss re-weighting \cite{cui2019class} and data re-sampling \cite{buda2018systematic} are two commonly-used strategies, which are employed at the training stage for a given labelled dataset. Active learning offers an alternative solution by selecting more samples from minority classes for labelling at the dataset construction stage. Aggarwal \etal \cite{aggarwal2020active} proposed to trigger class-balanced AL when the imbalance profile of the selected data meets some predefined condition, and the class-balanced AL is conducted by selecting the samples that are closest to under-represented classes and furthest away from any majority class in the feature space. Bengar \etal \cite{bengar2022class} perform class-balanced AL by minimizing the distance between the desired class distribution and the estimated distribution. Class-balanced strategy has been investigated for image-level ALOD, where the uncertainty of each box is weighted inversely by per-class occurrence \cite{brust2018active} or performance \cite{yamani2024active}. However, the weighting strategy obtains mixed performance depending on how the box-level scores are aggregated to generate the image-level score for selection \cite{brust2018active,yamani2024active}. In this work, we show that by performing selection at box-level and with proper pseudo labelling, our class-balanced sampling strategy is able to consistently outperform the non-balanced baseline.

\section{Method}
\begin{figure}[ht]

\centering
\includegraphics[width=0.9\linewidth]{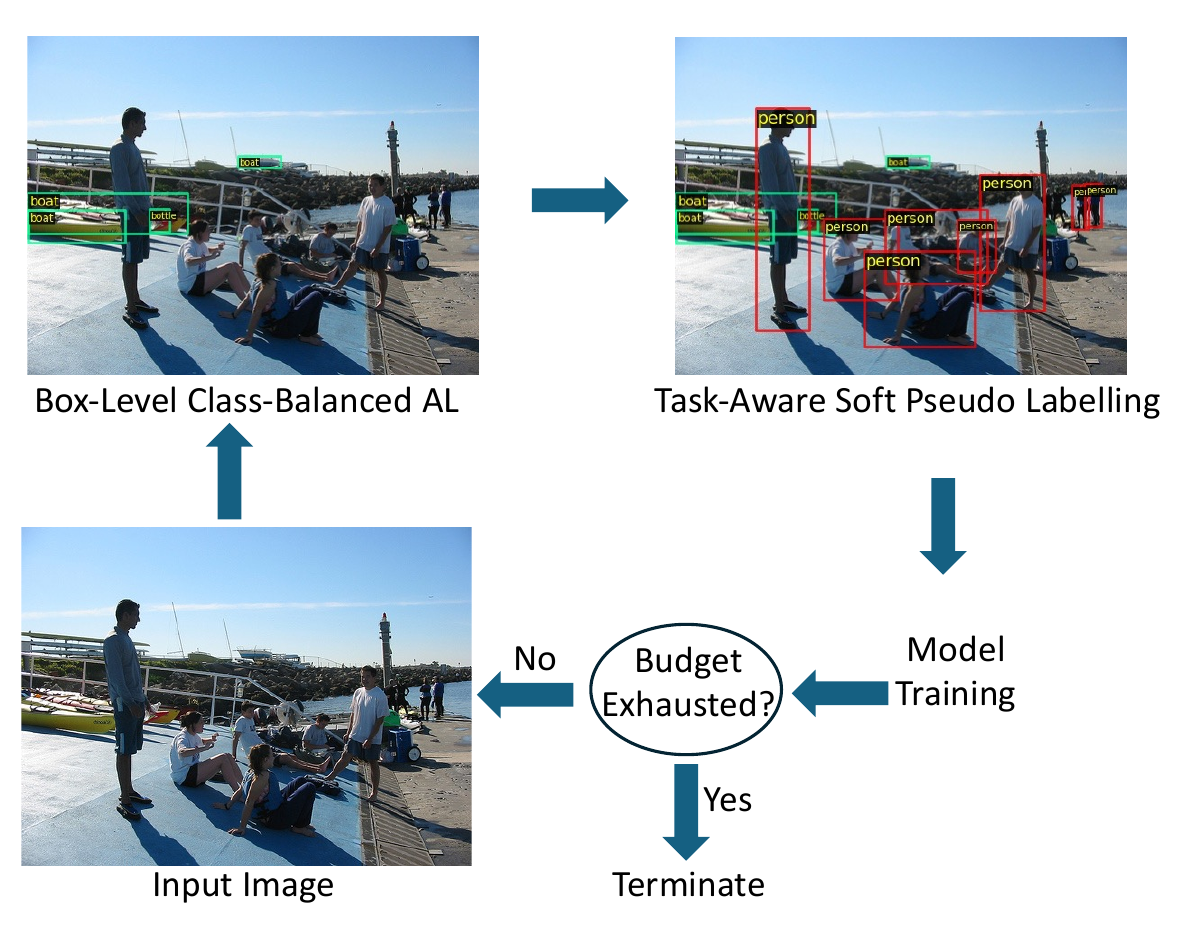}
\caption{The pipeline of the proposed method.}
\label{fig:system_diagram}

\end{figure}

Given an unlabelled dataset, we first train an object detector using a small random subset of fully-labelled images. The detector is then applied to the rest of unlabelled images and the detections with confidence score above certain threshold are considered as candidate objects. We propose a class-balanced AL strategy to select a subset of objects and query for their labels. For objects not selected by AL, we use those with reliable prediction as pseudo labels. The model is then retrained using all the GT labels and pseudo labels, and this process iterates until the annotation budget is exhausted. The pipeline of the proposed method is illustrated in Fig.~\ref{fig:system_diagram}. In this section, we detail our class-balanced selection strategy and task-aware soft pseudo labelling strategy. 

\subsection{Class-Balanced Sampling}
To bias the AL selection towards minority classes, we propose to weight the sample uncertainty inversely by the estimated class frequency: 
\begin{equation}
a(b) = u(b)e^{-\frac{p(\hat{l}_b)}{\sigma}},
\label{eq:ab}
\end{equation}
where $b$ denotes a candidate box, $u(b)$ is the uncertainty score of the box, $\hat{l}_b$ is the predicted label of the box,  $p(l), l=1,\ldots, K$ is the class probability distribution estimated from pseudo labels, and $\sigma$ is a hyperparameter that controls the balanced effect applied to different classes: smaller $\sigma$ favors more on the selection of minority classes.  The weighted uncertainty score $a(b)$ is sorted and the largest ones are selected. If $b$ is from a majority class, its uncertainty will be greatly down-weighted (as $p(\hat{l}_b)$ will be large), making it less likely to be selected; on the contrary, if $b$ is from a minority class, its uncertainty will be less down-weighted and thus more likely to be selected. 

There are various ways to measure the uncertainty of an unlabelled sample. For object detection, it is beneficial to consider the uncertainty in both class and box prediction  \cite{yu2022consistency,lyu2023box}, which can be measured based on the consistency of predictions over multiple augmentations. We adapt CALD \cite{yu2022consistency}, which is originally developed for image-level selection, to measure box-level uncertainty. Specifically, let $b$ denote a detection on the original image, and $b_m$ the corresponding prediction on augmented image $m$, the consistency score of $b$ over augmentations is measured by:
\begin{equation}
c(b) = \mathbb{E}_{m \in \set{M}} |\frac{1}{2}(s_b + s_{bm})(1-JS(b, b_m)) + IoU(b, b_m) - \beta|
\label{eq:cald_cons}
\end{equation}
where $\set{M}$ is the set of augmentations, $s_b$ and $s_{bm}$ is the confidence score for $b$ and $b_m$ respectively, $JS(b, b_m)$ is the Jensen-Shannon (JS) divergence between the predicted class distribution of $b$ and $b_m$, and $IoU(b, b_m)$ is the Intersection-over-Union (IoU) between the predicted box of $b$ and $b_m$, and $\beta$ is set to 1.3 following \cite{yu2022consistency}. As confidence score, JS divergence and IoU are all bounded by [0,1], the value of $c(b)$ is bounded by $[0, c_m]$, where $c_m = max(\beta, 2-\beta)$. We propose the following uncertainty measure based on the consistency score:
\begin{equation}
u(b)=c_m - c(b),
\label{eq:ub}
\end{equation}
which is plugged into Eq.~(\ref{eq:ab}) to compute $a(b)$.

\subsection{Task-Aware Soft Pseudo Labelling}
With box-level selection and annotation, a training image may be sparsely annotated, \ie, objects selected by AL are provided with class label and bounding box coordinates, while those unselected regions are left unlabelled. This can cause a foreground anchor or proposal being assigned as negative sample and harm model performance. We propose a task-aware soft pseudo labelling strategy to address this problem. There are two key ingredients in the proposed strategy:

\myparagraph{Task-aware pseudo labelling} Object detection involves two tasks: predicting 1) the class label and 2) the bounding box coordinates for the object of interest. Previous work \cite{desai2020towards} simply used predictions with high confidence score as pseudo labels, which ignores the fact that detection with high confidence score may not have accurate box prediction. We propose to use different pseudo labels to supervise the two tasks. For classification branch, the predicted confidence score is used as criterion, and predictions with confidence score above threshold $\tau^{cls}_0$ are used to supervise the classification branch. For localization branch, we compute the consistency of the box prediction over multiple augmentations in a way similar to Eq.~(\ref{eq:cald_cons}):
\begin{equation}
cons(b) = \mathbb{E}_{m \in \set{M}} IoU(b, b_m).
\label{eq:cons}
\end{equation}
Predictions with consistency score above threshold $\tau^{box}_0$ are used to supervise the localization branch.

\myparagraph{Soft pseudo labelling} To obtain pseudo labels of high precision, a large value is set for $\tau^{cls}_0$ and $\tau^{box}_0$. However, detections with confidence score or box consistency score below that threshold may still indicate the presence of GT objects, and treating such area as background will provide incorrect supervision signal for model training. To utilize these detections during training, we propose to assign a weight to each GT or pseudo box:
\begin{equation}
\small
    w^{cls}_b = \begin{cases}
             1  & \text{GT box or pseudo box with  $conf(b) >= \tau^{cls}_0$} \\
        conf(b) & \text{pseudo box with  $\tau^{cls}_1 < conf(b) < \tau^{cls}_0$} \\  
             0  & \text{pseudo box with  $conf(b) <= \tau^{cls}_1$} \\ 
    \end{cases}
\label{eq:w_cls}
\end{equation}
where $conf(b)$ is the confidence score of the predicted box. The box weight $w^{box}_b$ is similarly defined using the box consistency score (Eq.~(\ref{eq:cons})):
\begin{equation}
\small
    w^{box}_b = \begin{cases}
             1  & \text{GT box or pseudo box with  $cons(b) >= \tau^{box}_0$} \\
        cons(b) & \text{pseudo box with  $\tau^{box}_1 < cons(b) < \tau^{box}_0$} \\  
             0  & \text{pseudo box with  $cons(b) <= \tau^{box}_1$} \\ 
    \end{cases}
\label{eq:w_box}
\end{equation}

During training, an anchor or proposal is considered as a positive sample if it meets some IoU criterion with a GT or pseudo box. Its classification and localization loss are then weighted by the weights of the corresponding GT/pseudo box. Note that a positive sample may have different weights for classification and localization. The per-sample loss is summed up and normalized to get the final loss:
\begin{equation}
\begin{split}
\L  & = \frac{1}{|\set{P}_{cls}|} \sum_{x \in \set{P}_{cls}} w^{cls}_x \L_{cls}(cls(x), t^{cls}_x)  \\
    & + \frac{1}{|\set{N}_{cls}|} \sum_{x \in \set{N}_{cls}} \L_{cls}(cls(x), t^{cls}_x) \\
    & + \frac{1}{|\set{P}_{reg}|} \sum_{x \in \set{P}_{reg}} w^{box}_x \L_{loc}(reg(x), t^{reg}_x),
\end{split}
\end{equation}
where $\set{P}_{cls/reg}$ is the set of positive samples (anchors or proposals) for classification/localization branch, and $\set{N}_{cls}$ is the set of negative samples for classification, $cls(x)$ is the classification branch prediction for sample $x$,  $reg(x)$ is the regression (localization) branch prediction for sample $x$, and $t_{cls/reg}(x)$ is the training target for $x$.

\section{Experiments}
\subsection{Experimental Setup}
\myparagraph{Dataset} VOC0712 \cite{everingham2010pascal} consists of images from 20 object classes. Following the setting of previous works \cite{yoo2019learning,wu2022entropy}, {\tt trainval'07} (5,011 images) and {\tt trainval'12} (11,540 images) are combined to make a super-set {\tt trainval'0712} (16,551 images) and used as the initial unlabelled pool. The performance is reported on {\tt test'07} (4,952 images). 

\myparagraph{Implementation details} We use the open-source MMDetection \cite{mmdetection} framework and evaluate with one-stage detector RetinaNet \cite{lin2017focal} as well as two-stage detector Faster R-CNN \cite{ren2015faster}. Both detectors use feature pyramid networks \cite{lin2017feature} on top of ResNet-50 \cite{he2016deep} as feature extractor. The hyperparameters are set as below: total training epochs = 26, initial learning rate = 1e-3 for RetinaNet and 5e-3 for Faster R-CNN, which is decayed by 0.1 after 20 epochs, and batch size = 2. During active learning, all methods start from the same initial pool containing 1000 randomly selected fully-annotated images, and another 2k, 4k and 8k boxes are selected for labelling in subsequent cycles using different AL strategies. The parameters for our method are set as below: $\sigma$ in Eq.~(\ref{eq:ab}) is set to 1, $\tau^{cls}_0$ and $\tau^{cls}_1$ in Eq.~(\ref{eq:w_cls}) are set to 0.8 and 0.1, $\tau^{box}_0$ and $\tau^{box}_1$ in Eq.~(\ref{eq:w_box}) are set to 0.7 and 0.3. We report the mean and standard deviation of three runs that use different seeds to generate the initial pool.


\subsection{Quantitative Results}
\begin{figure}[ht]
\centering
\begin{subfigure}{0.23\textwidth}
	\includegraphics[width=\linewidth]{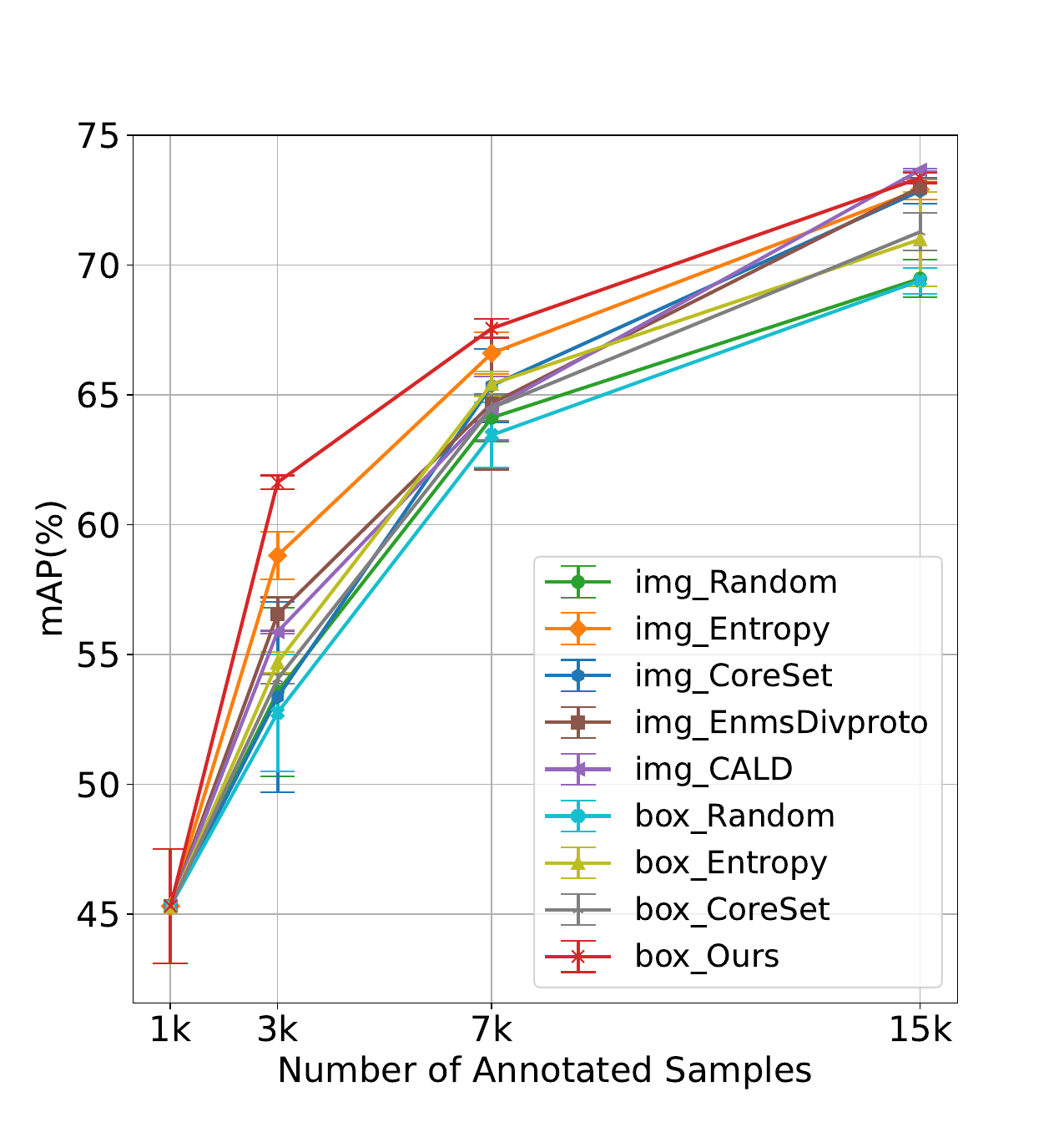}
	\caption{}
	\label{fig:al_retinanet_voc0712}
\end{subfigure}
\begin{subfigure}{0.23\textwidth}
	\includegraphics[width=\linewidth]{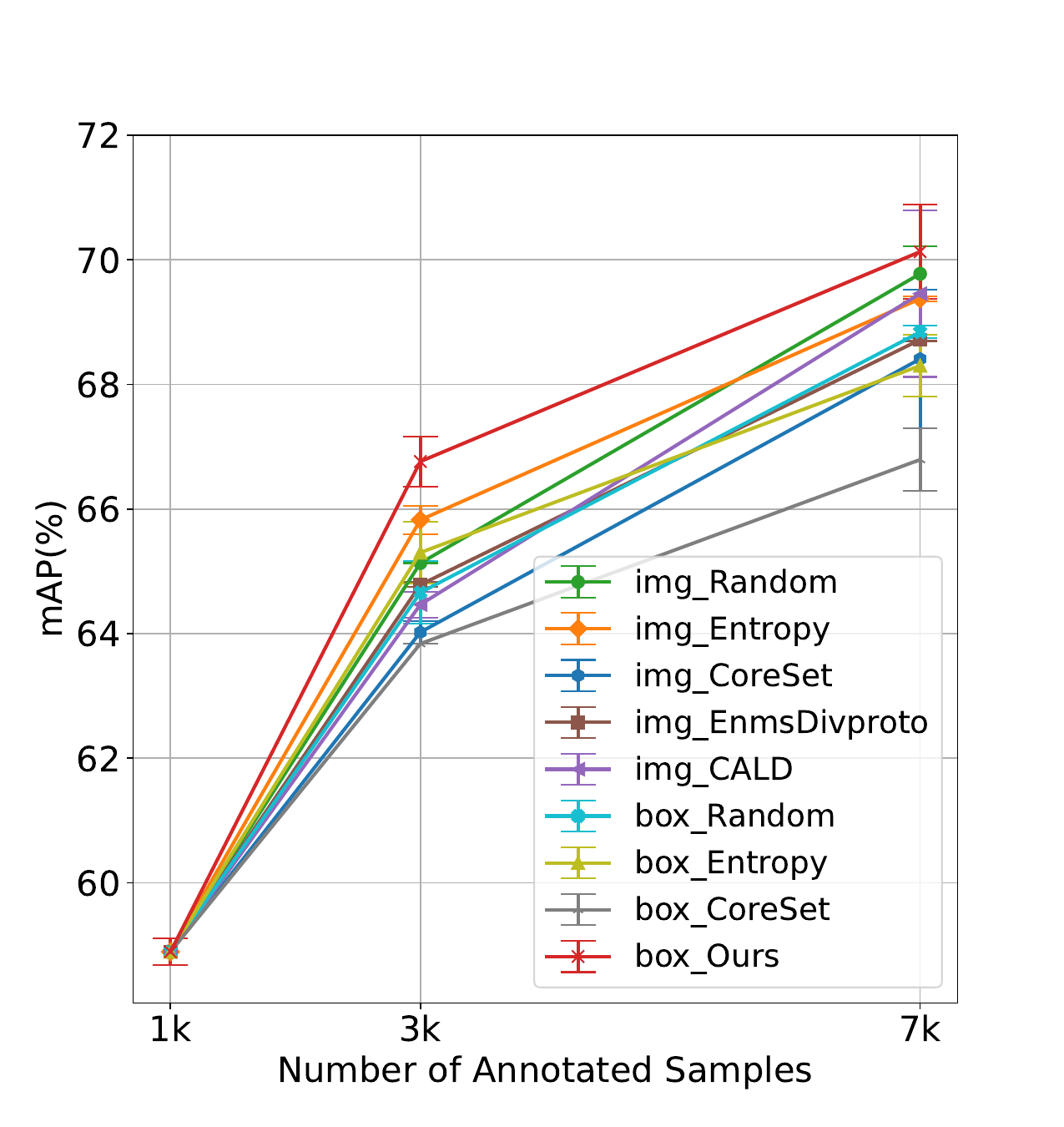}
	\caption{}
	\label{fig:al_faster_rcnn_voc0712}
\end{subfigure}
\caption{Active learning results on VOC0712. (a) RetinaNet. (b) Faster R-CNN.}
\label{fig:al_results}
\end{figure}

\myparagraph{Benchmarking results} We benchmark with image-level methods (Random, Entropy, CoreSet \cite{sener2017active}, EnmsDivproto \cite{wu2022entropy} and CALD \cite{yu2022consistency}) as well as box-level methods (Random, Entropy, CoreSet). We report AL results on RetinaNet  and Faster R-CNN in Fig.~\ref{fig:al_results}. We observe that our method is able to consistently outperform competing methods at different budgets. The gain is most significant when the budget is low, \eg, at the second batch (budget=3k), our method beat the second best method by 2.8\% on RetinaNet, and 0.9\% on Faster R-CNN. We also observe that on Faster R-CNN, state-of-the-art image-level methods, \eg, EnmsDivproto and CALD, cannot beat the Random baseline  when the annotation cost is measured by the number of boxes, highlighting the importance of using more realistic cost measurement when benchmarking different methods.

\myparagraph{Ablation studies}
We report the ablation studies on the design choices of our method in Table~\ref{tab:ablation_study}. The proposed class-balanced sampling is able to improve upon the non-balanced baseline (uncertainty sampling using the same uncertainty measure as used in class-balanced sampling) by 0.5$\sim$0.7\% mAP. Hard pseudo labelling can only bring marginal improvement (\ie, $\sim$0.3\% mAP), indicating that using only predictions of high confidence as pseudo labels and ignoring others is sub-optimal as it can cause foreground region being mistakenly treated as background. The proposed soft pseudo labelling brings significant gain (up to 3.8\% mAP) compared to hard pseudo labelling, suggesting that using low confidence prediction as pseudo labels and down-weight their contribution to loss computation accordingly is effective in handling sparsely labelled images for box-level AL.

\begin{table*}
\centering
\caption{Ablation studies on the design choices of our method. U: uncertainty sampling (using Eq.~(\ref{eq:ub})), CB represents class-balanced sampling (using Eq.~(\ref{eq:ab})), HPL: hard pseudo labelling ($\tau^{cls}$ = 0.8), THPL: task-aware hard pseudo labelling ($\tau^{cls}=0.8$ and $\tau^{box}=0.7$), TSPL: task-aware soft pseudo labelling ($\tau^{cls}_0=0.8$, $\tau^{cls}_1=0.1$, $\tau^{box}_0=0.7$, $\tau^{box}_1=0.3$). Results are reported on VOC0712 with RetinaNet. Numbers in brackets are standard deviation of three runs.}
\begin{tabular}{ccccccccc}
\toprule
U          & CB         & HPL           &  THPL  & TSPL            & Batch 2      & Batch 3 & Batch 4\\
\midrule
\checkmark &            &   &    &  & 57.13(1.03) & 64.26(3.29) & 72.54(0.20)\\
\checkmark & \checkmark &   &    &   & 57.76(1.07) & 64.93(2.14) & 73.25(0.78)\\
\checkmark & \checkmark & \checkmark  &    &   & 57.72(0.79) & 65.00(1.22) & 73.31(0.63)\\
\checkmark & \checkmark &  &  \checkmark  &   & 58.06(0.49) & 65.21(0.42) & 73.34(0.19)\\
\checkmark & \checkmark &  &  & \checkmark &  \textbf{61.90(0.88)} & \textbf{67.92(1.87)} & \textbf{73.56(0.32)}\\
\bottomrule
\end{tabular}
\label{tab:ablation_study}
\end{table*}

\subsection{Qualitative Results}
\begin{figure*}[ht]
\centering
\includegraphics[width=1.0\linewidth]{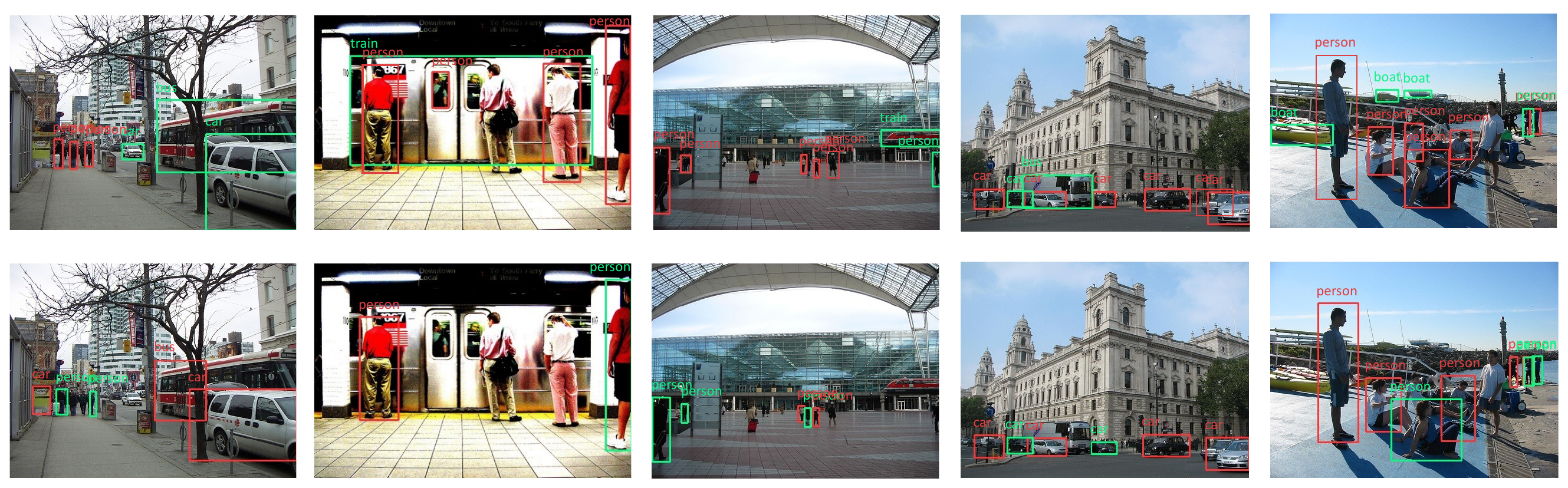}
\caption{Qualitative results of RetinaNet on VOC0712. Top row: boxes selected by our class-balanced sampling method (U+CB in Table~\ref{tab:ablation_study}). Bottom row: boxes selected by uncertainty sampling (U in Table~\ref{tab:ablation_study}). Boxes in cyan are those selected for ground truth labelling by AL, and boxes in red are pseudo labels. Our method is able to select more samples from minority classes (\eg, bus, train, boat) for labelling, and majority classes can be reasonably labelled by pseudo labels; while other methods select most boxes from the majority class (\eg, person) for labelling and minority classes cannot be well compensated by pseudo labelling.}
\label{fig:qualitative_results}
\end{figure*}

We visualize the samples selected by various methods in Fig.~\ref{fig:qualitative_results}. Our method selects more samples from minority classes (\eg, bus, train, boat) for labelling. As current detector is already able to perform well on majority classes, unlabelled objects from majority classes can be well handled by pseudo labelling. In contrast, other AL methods select most boxes from the majority class (\eg, person) for labelling. As the detector does not perform well on minority classes, it cannot provide high quality pseudo labels for regions that contain minority objects and these regions are incorrectly treated as background during training. 

\section{Conclusion}
In this work, we propose a class-balanced sampling strategy for box-level ALOD. Our method weights the  uncertainty of each object inversely by the estimated class frequency to bias the AL selection towards minority classes. We complement sparsely labelled images with pseudo labels and propose a task-aware soft pseudo labelling strategy, where different criteria are used to filter the pseudo labels for the classification and localization branches, and the per-sample loss is weighted by the reliability of the assigned GT/pseudo box. We evaluate our method with both one-stage and two-stage detectors on VOC0712 dataset. Experimental results show that the proposed class-balanced sampling strategy is able to select more objects from minority classes for labelling, and unlabelled objects in sparsely labelled images can be well handled by the proposed soft pseudo labelling strategy. We benchmark our method with other image-level and box-level methods, and demonstrate it achieves state-of-the-art performance.


\bibliographystyle{IEEEbib}
\bibliography{references}

\end{document}